\documentclass[10pt,twocolumn,letterpaper]{article}

\usepackage{iccv}              

\definecolor{iccvblue}{rgb}{0.21,0.49,0.74}
\usepackage[pagebackref,breaklinks,colorlinks,allcolors=iccvblue]{hyperref}
\usepackage{multirow}
\usepackage{tabularx}
\usepackage{threeparttable}
\usepackage{algorithm}
\usepackage{algpseudocode}
\usepackage[accsupp]{axessibility}

\def\paperID{4262} 
\def\confName{ICCV}
\def\confYear{2025}

\title{HoliTracer: Holistic Vectorization of Geographic Objects from Large-Size Remote Sensing Imagery}

\author{
  Yu Wang, Bo Dang, Wanchun Li, Wei Chen, Yansheng Li\textsuperscript{*} \\
  \vspace{5px}
  School of Remote Sensing and Information Engineering, Wuhan University \\
  \tt\small \{wangfaye,bodang,wanchun.li,weichenrs,yansheng.li\}@whu.edu.cn
}
\begin{document}

\twocolumn[{ \renewcommand\twocolumn[1][]{#1}
    \maketitle
    \begin{center}
        \centering
        \includegraphics[width=\textwidth]{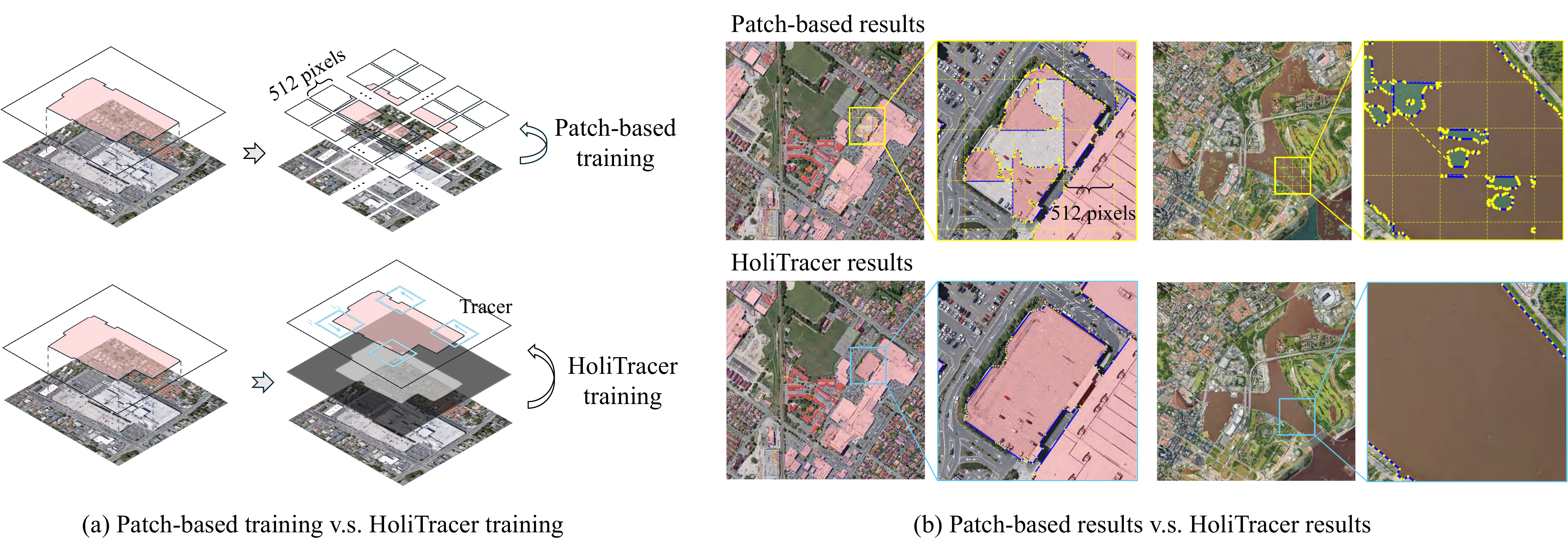}
        \vspace{-15pt}
        \captionof{figure}{\textbf{Existing Patch-based method \cite{xu2023hisup} v.s. our HoliTracer.}. 
           Existing methods adopt a patch-based approach during training and inference, leading to context loss and fragmented vector results. 
           In contrast, our HoliTracer perceives context while directly tracing entire objects, yielding superior vector results.
           }
        \label{fig:motivation}
    \end{center}
}]

\begingroup
\renewcommand\thefootnote{\fnsymbol{footnote}}
\footnotetext[1]{Corresponding author.}
\endgroup

\begin{abstract}
With the increasing resolution of remote sensing imagery (RSI), large-size RSI has emerged as a vital data source for high-precision vector mapping of geographic objects.  
Existing methods are typically constrained to processing small image patches, which often leads to the loss of contextual information and produces fragmented vector outputs.  
To address these, this paper introduces \textbf{HoliTracer}, the first framework designed to holistically extract vectorized geographic objects from large-size RSI.  
In HoliTracer, we enhance segmentation of large-size RSI using the Context Attention Net (CAN), which employs a local-to-global attention mechanism to capture contextual dependencies. 
Furthermore, we achieve holistic vectorization through a robust pipeline that leverages the Mask Contour Reformer (MCR) to reconstruct polygons and the Polygon Sequence Tracer (PST) to trace vertices.  
Extensive experiments on large-size RSI datasets, including buildings, water bodies, and roads, demonstrate that HoliTracer outperforms state-of-the-art methods.  
Our code and data are available in \href{https://github.com/vvangfaye/HoliTracer}{github.com/vvangfaye/HoliTracer}
\end{abstract}
\vspace{-10px}
\section{Introduction}
\label{sec:intro}
Vector maps provide precise representations of the Earth’s surface and are essential for various downstream tasks, such as navigation and urban planning~\cite{bongiorno2021vector,zheng2023spatial}. 
Automatic extraction of accurate vector maps from remote sensing imagery (RSI) has emerged as a cost-effective approach, achieving significant progress in recent years~\cite{li2024review,chen2022road}.  

With advancements in remote sensing technology, the resolution of RSI continues to improve, resulting in increasingly large images that require interpretation~\cite{alvarez2021uav, roy2017satellite, li2025meet}.  
However, existing vector mapping methods~\cite{girard2021polygonal,xu2023hisup,wang2023regularized,yang2024univecmapper} are typically designed for small image patches and struggle to process such large-size imagery, as illustrated in Fig.~\ref{fig:motivation}.  
Specifically, due to computational constraints, most vectorization algorithms are limited to processing image inputs of $512 \times 512$ pixels~\cite{girard2021polygonal,yang2024univecmapper}.  
When applied to large-size imagery exceeding $10,000 \times 10,000$ pixels, these methods resort to a simplistic patch-based strategy—cropping the image, processing patches independently, and merging the results~\cite{xu2023hisup}.  
This strategy, however, introduces significant challenges, referred to here as the ``large-size challenge," as depicted in Fig.~\ref{fig:motivation}b.  
On one hand, the patch-based strategy discards critical contextual information in large-size RSI, hindering the model’s ability to accurately distinguish objects requiring broader context. For instance, buildings may be confused with parking lots without sufficient surrounding information, as shown in Fig.~\ref{fig:motivation}b.  
On the other hand, vector outputs derived from individual patches often exhibit fragmentation at patch boundaries, compromising the geometric integrity of the results, as illustrated in Fig.~\ref{fig:motivation}b.  

In addition to the ``large-size challenge" posed by patch-based methods, geographic objects in large-size RSI exhibit significant scale variations across different categories.  
For example, continuous water bodies typically require more vector points for representation than scattered buildings.  
This scale variation exacerbates the fragmentation issue and poses challenges for achieving a unified representation across diverse object categories.  
While existing research has primarily focused on single-category extraction, such as buildings~\cite{li2024review} or roads~\cite{chen2022road}, a few frameworks have been proposed for unified geographic object extraction~\cite{wang2023regularized,yang2023topdig}.  
Nevertheless, these methods remain constrained by their reliance on patch-based processing, failing to address the large-size challenge effectively.  
Consequently, there is a pressing need to develop a unified vectorization approach capable of directly handling large-size RSI.

To address these challenges, this paper proposes a framework called HoliTracer for holistic vector extraction directly from large-size RSI.  
To enable holistic vectorization under limited computational resources, HoliTracer draws inspiration from segmentation-based methods~\cite{wei2019toward}, which adopt a two-stage process of segmentation followed by vectorization.
HoliTracer first performs segmentation on large-size RSI, leveraging the fact that pixel-level extraction preserves feature completeness without requiring post-processing while effectively capturing contextual information. 
Subsequently, based on these complete segmentations, HoliTracer traces the contours of each segmentation polygon to generate the final vector results.  
More specifically, HoliTracer employs a Context Attention Network (CAN) to capture information using a local-to-global attention mechanism within large-size RSI. By adaptively integrating this information, CAN achieves more complete segmentation compared to patch-based methods.  
Thereafter, to derive vector results from the segmentation mask, we introduce a robust vectorization pipeline leveraging the Mask Contour Reformer (MCR) and the Polygon Sequence Tracer (PST).  
MCR reconstructs irregular polygon contours while ensuring alignment with ground truth polygons through a bidirectional matching mechanism.  
The reconstructed polygons are then processed by PST for polygon refinement and vertex identification, yielding precise vector representations.  
Thanks to the robustness of the MCR algorithm and PST's sequence tracing strategy, HoliTracer effectively handles objects across diverse categories and scales.  
We conducted comparative experiments on multiple large-size datasets featuring various geographic objects, including buildings, water bodies, and roads.  
Extensive experiments demonstrate that our method significantly outperforms existing patch-based approaches in vectorizing large-size RSI.  
In summary, our contributions are as follows:
\begin{itemize}
  \item To the best of our knowledge, HoliTracer is the first method designed for large-size RSI vectorization, holistically extracting diverse geographic objects.
  \item We propose CAN, using a local-to-global attention mechanism to enhance segmentation and address context loss in patch-based methods.
  \item We design a pipeline with MCR and PST for precise polygon reconstruction and vertex tracing across varied objects in large-size RSI.
  \item Experiments on large-size RSI datasets of buildings, water bodies, and roads show HoliTracer outperforms existing state-of-the-art methods.
\end{itemize}

\begin{figure*}
  \begin{center}
    \includegraphics[width=1\textwidth]{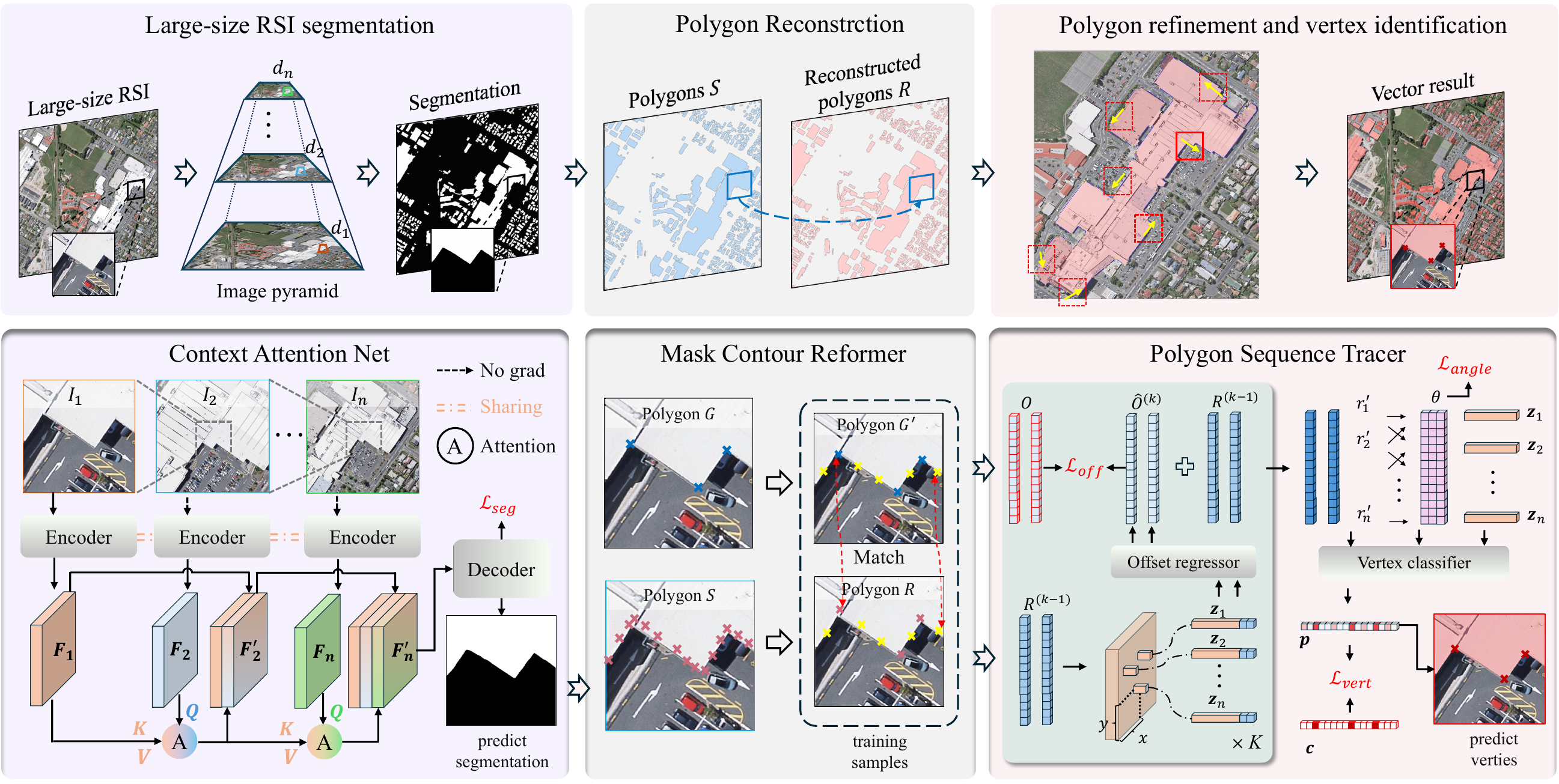}
  \end{center}
  \vspace{-12pt}
     \caption{
      The overall framework of HoliTracer. HoliTracer includes Context Attention Net (CAN) for context understanding from large-size RSI, Mask Contour Reformer (MCR) for polygon contours reconstruction, and Polygon Sequence Tracer (PST) for polygon refinement and vertex identification. With the proposed pipeline, HoliTracer can directly extract diverse geographic objects from large-size RSI.
      }
  \label{fig:framework}
\end{figure*}
\section{Related Work}
\subsection{Vectorization of Remote Sensing Imagery}
\textbf{Segmentation-based Methods}.  
Segmentation-based methods often employ a two-step strategy to vectorize segmentation masks~\cite{zhou2020bt,wei2019toward}.  
These approaches first generate irregular binary masks and then derive vectors by simplifying raster maps.
For instance, Wei et al.~\cite{wei2019toward} extracted building polygon from binary masks using the DP algorithm~\cite{douglas1973algorithms}, followed by post-processing to refine the polygons.  
Additionally, some methods integrate vectorization with segmentation through multi-task learning~\cite{hatamizadeh2020end,girard2021polygonal,xu2023hisup}, achieving improved performance by jointly optimizing both tasks.  
These approaches learn building masks alongside auxiliary information, such as vector fields and  corner locations, to construct vector mapping structures.  
Other studies emphasize vectorization techniques, such as the GGT model~\cite{belli2019image}, which leverages a self-attention mechanism to generate vectorized roads from binary masks~\cite{xu2022csboundary}.

\textbf{Contour-based Methods}.  
Contour-based methods directly extract vector topologies from input images.  
Typically, initial contours are obtained using object detectors or segmentation techniques, followed by refinement to produce final vector results.  
Early approaches often relied on fixed initial contour templates for each instance~\cite{ling2019fast,wei2021graph,peng2020deep,liang2020polytransform,zhang2025vectorllm}.  
DeepSnake~\cite{peng2020deep} transforms object bounding boxes into octagons, achieving higher accuracy.  
Recent methods replace handcrafted initial templates with coarse contours to enhance refinement performance~\cite{zhang2022e2ec,zhu2022sharpcontour,huang2024instance}.  
Zhang et al.~\cite{zhang2022e2ec}, for instance, introduced the E2EC workflow, which progressively refines coarse segmentation boundaries, outperforming prior approaches.

\textbf{Graph-based Methods}.  
Graph-based methods generate vector topologies by predicting nodes and their connections.  
One subset focuses on road graph delineation by tracing sequential nodes~\cite{bastani2018roadtracer,tan2020vecroad,xu2022rngdet}.  
Another subset first identifies target nodes and then connects potential node pairs to form topological graphs~\cite{zhu2021adaptive,zhang2019ppgnet,xu2022csboundary,zorzi2022polyworld}.  
Recently, graph-based methods have been developed to uniformly extract multiple categories of geographic objects, such as buildings, roads, and water bodies~\cite{wang2023regularized,yang2023topdig,yang2024univecmapper}.  
UniVec~\cite{yang2024univecmapper}, for example, is a universal model that extracts directional topological graphs from remote sensing images across object classes, utilizing a topology-focused node detector and a perturbed graph supervision strategy.

While these methods have shown promising results in vectorizing geographic objects, they are generally designed for patch-sized RSI and struggle to interpret large-size RSI.  
Moreover, most approaches target single-category geographic objects, limiting their ability to handle multi-category extraction.  
In contrast, our HoliTracer is specifically tailored for large-size RSI, enabling holistic extraction of multiple geographic object categories from such imagery.
The holistic object we refer to is a polygon that contains complete information about a geographic object, enabling a better representation of the object's shape and position.

\subsection{Large-size RSI Interpretation}
Recently, large-size RSI interpretation has gained increasing attention, encompassing tasks such as semantic segmentation~\cite{liu2024ultra,ding2020semantic} and object detection~\cite{li2024learning,li2025star}.  
For large-size semantic segmentation, TS-MTA~\cite{ding2020semantic} trains models at local and global scales separately to develop multi-view perception capabilities.  
MFVNet~\cite{li2023mfvnet} employs a multi-view segmentation strategy, processing remote sensing images at different scales and dynamically fusing the results.  
LCF-ALE~\cite{liu2024ultra} introduces a local context fusion and enhancement module to handle large-size RSI.  
In object detection, methods like HBDNet~\cite{li2024learning} address specific challenges, such as detecting bridges across scales and automatically integrating multi-scale results.  
However, for the more complex task of vector mapping, no existing research specifically tackles large-size vectorization.
\section{HoliTracer}
The overall framework of HoliTracer is illustrated in Fig.~\ref{fig:framework}. 
HoliTracer incorporates the CAN to extract context information using a local-to-global attention mechanism from large-size RSI.  
Building upon this, to generate more robust and trainable polygon inputs, HoliTracer includes the MCR, which reconstructs polygons and aligns them with ground truth to produce training samples.  
Finally, the PST is introduced as a sequence tracing module to refine polygons and identify vertices, yielding precise vector representations.  
The following sections provide detailed descriptions of these three components.
\subsection{Context Attention Net}

\textbf{Multi-Scale Image Pyramid}. 
Unlike current methods that directly slice the original large-size RSI, CAN slices a multi-scale image pyramid to enable the model to leverage a local-to-global attention mechanism. 
Specifically, given a large-size RSI, we first apply different downsampling rates $d_1, d_2, \ldots, d_n$ to construct a multi-scale image pyramid of the original image. 
The bottom layer of the pyramid has a downsampling rate of 1, with rates increasing from bottom to top such that $d_1 < d_2 < \ldots < d_n$. 
As shown in Fig.~\ref{fig:framework}, we use a sliding window of uniform size to slice the image pyramid. 
This results in multiple image patches of the same size: ($I_1, I_2, \ldots, I_n$), where the range of the next layer $I_{n-1}$ is centered within the previous layer $I_n$. 
These images collectively provide multi-scale observations of $I_1$, supporting the local-to-global attention mechanism for understanding context information.

\textbf{Context Attention}. 
For each set of pyramid slices, images from different scales are processed by a unified image encoder: $F_n = \text{Encoder}(I_n)$, where $F_n$ denotes the feature map of the $n$-th scale image. The encoder is Swin-L~\cite{liu2021swin} with Skysense~\cite{guo2024skysense} weights.
Notably, only the bottom layer encoder is trained, with upper layers sharing its weights.

After extracting features across scales, we employ a Context Attention approach to adaptively fuse the bottom layer feature \(\mathbf{F}_1\) with each of \(\mathbf{F}_2, \ldots, \mathbf{F}_n\) using a attention mechanism. 
Specifically, for the \(n\)-th scale image, it is fused with the bottom layer feature through attention:
\begin{equation}
\mathbf{F}'_n = \text{Softmax} \left( \frac{\mathbf{Q} \mathbf{K}_n^T}{\sqrt{d}} \right) \cdot \mathbf{V}_n,
\end{equation}
where \(\mathbf{Q} = \mathbf{F}_1\) and \(\mathbf{K}_n = \mathbf{V}_n = \mathbf{F}_n\). 
This approach enables features at each scale to leverage both local and global information via the attention mechanism. 
The fused features are then concatenated and fed into the semantic segmentation decoder:
\begin{equation}
\mathbf{F} = \text{Concat} \left[ \mathbf{F}_1, \mathbf{F}'_2, \ldots, \mathbf{F}'_n \right].
\end{equation}

\textbf{Segmentation Decoder and Loss Function}. 
We adopt the widely used UperNet segmentation head~\cite{xiao2018unified} for its strong multi-scale fusion capability. 
Additionally, we train CAN using a cross-entropy loss function, defined as:
\begin{equation}
\mathcal{L}_{\text{seg}} = -\frac{1}{N} \sum_{i=1}^{N} \sum_{c=1}^{C} y_{i,c} \log(\hat{y}_{i,c}),
\end{equation}
where \(N\) is the number of pixels, \(C\) is the number of classes, \(y_{i,c}\) is the ground truth label, and \(\hat{y}_{i,c}\) is the predicted probability of class \(c\) at pixel \(i\).

\subsection{Mask Contour Reformer}

\begin{figure}
  \begin{center}
    \includegraphics[width=0.48\textwidth]{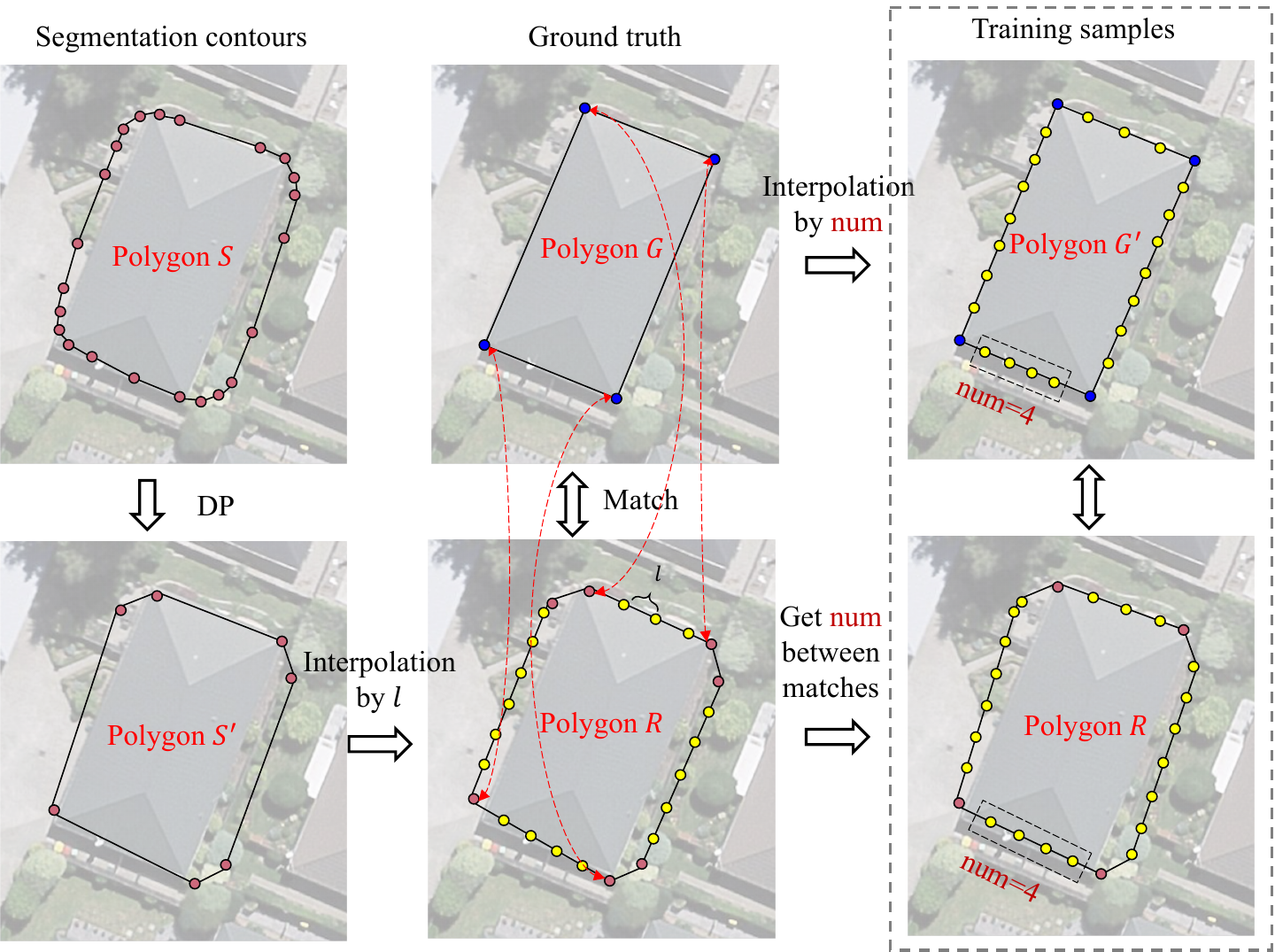}
  \end{center}
  \vspace{-12pt}
     \caption{The pipeline of Mask Contour Reformer.}
  \vspace{-12pt}
  \label{fig:match}
\end{figure}
The pipeline of MCR is illustrated in Fig.~\ref{fig:match}.  
From segmentation results, polygon contours are extracted as a set \( S = \{s_1, s_2, \dots, s_n\} \) using TC89-KCOS~\cite{teh1989detection}, where each \( s_i = (x_i, y_i) \) represents a 2D coordinate.  
These points are typically dense and irregular, making direct refinement and correspondence with ground truth polygons challenging.  
To address this, MCR adopts a simplify-then-reconstruct approach, starting with the Douglas-Peucker algorithm~\cite{douglas1973algorithms} applied to \( S \) with a tolerance parameter \( \epsilon \).  
This yields a simplified polygon \( S' = [s'_1, s'_2, \dots, s'_m] \) (where \( m \leq n \) and \( S' \subset S \)).  
Subsequently, MCR regularizes the polygon by interpolating points along each edge of \( S' \) at a fixed distance \( l \).  
The reconstructed segmentation polygon is obtained as \( R = [r_1, r_2, \dots, r_N] \), where \( N = m + \sum_{i=1}^m K_i \), and \( K_i = \lfloor \|s'_{i+1} - s'_i\| / l \rfloor \) for each edge.

During the training stage, MCR aligns \( R \) with the ground truth polygon \( G = [g_1, g_2, \dots, g_M] \) using a bidirectional matching mechanism.  
For each \( g_j \in G \), the closest point in \( R \) is identified as \( r_{k_j} = \arg\min_{r \in R} \| r - g_j \| \), where \( \| r - g_j \| = \sqrt{(r_x - g_{j,x})^2 + (r_y - g_{j,y})^2} \).  
Unique matched points \( \{ r_{k_j} \mid j = 1, 2, \dots, M \} \) are designated as vertices, with indices \( i_1 < i_2 < \dots < i_P \) (where \( P \leq M \) if multiple \( g_j \) map to the same \( r_k \)), assumed to correspond to \( g_1, g_2, \dots, g_P \) in order.  
Remaining points in \( R \) are classified as non-vertices, and labels are assigned as \( c_i = 1 \) if \( r_i \) is a vertex, otherwise \( c_i = 0 \), forming \( C = [c_1, c_2, \dots, c_N] \) for categorical constraints.  
For each pair of consecutive vertices \( r_{i_k} \) and \( r_{i_{k+1}} \), the number of non-vertices is \( n_k = i_{k+1} - i_k - 1 \) (with \( i_{M+1} = i_1 \)), and the corresponding ground truth edge \( (g_k, g_{k+1}) \) is interpolated with \( n_k \) points.  
Concatenating all points produces the reconstructed ground truth \( G' \), with
\begin{equation}
  |G'| = M + \sum_{k=1}^M n_k = N = |R|.
\end{equation}
This ensures a one-to-one correspondence with \( R \).

In the prediction stage, MCR applies only the reconstruction process to segmentation results, outputting \( R \) without matching or ground truth interpolation.  
\textit{The complete algorithm is detailed in supplementary Algorithm~\ref{alg:mcr}.}

\subsection{Polygon Sequence Tracer}

Given a reconstructed polygon \( R = [r_1, r_2, \dots, r_n] \), the Polygon Sequence Tracer (PST) processes the polygon as a sequence, performing offset refinement and vertex classification for each point, as illustrated in Fig.~\ref{fig:framework}.  
Recognizing the importance of angles between points in vertex detection (where smaller angles indicate a higher likelihood of vertices), PST incorporates angle feature extraction and an angle penalty loss.

Specifically, the process begins with image feature extraction utilizing the Encoder from the segmentation stage, whose weights are frozen during training. For each point \( r_i \), point features are extracted from the image feature around \( (x_i, y_i) \), yielding a feature vector \( \mathbf{z}_i \). 
This vector is concatenated with the point's coordinates, forming the input \( [\mathbf{z}_i, x_i, y_i] \), which is fed into a transformer-based offset regressor. 
The regressor predicts an offset correction \( \hat{o}_i = (\Delta x_i, \Delta y_i) \) for each point, adjusting its position to \( r'_i = r_i + \hat{o}_i \). 
Inspired by DeepSnake \cite{peng2020deep}, multiple iterations of offset regression are applied, where the corrected sequence \( R^{(k)} \) at iteration \( k \) is updated as:
\begin{equation}
  R^{(k)} = R^{(k-1)} + \hat{O}^{(k)}, \quad \hat{O}^{(k)} = [\hat{o}_1^{(k)}, \hat{o}_2^{(k)}, \dots, \hat{o}_n^{(k)}]
\end{equation}
This iterative refinement captures subtle movements, improving the polygon’s alignment with the underlying structure.

After obtaining the corrected sequence \( R' = [r'_1, r'_2, \dots, r'_n] \), PST computes angles at each point \( r'_i \) using points at distances \( s = 1, 2, 3 \) (i.e., \( r'_{i-1}, r'_{i+1} \), \( r'_{i-2}, r'_{i+2} \), \( r'_{i-3}, r'_{i+3} \)) with periodic boundary conditions for a closed polygon (\( r'_i = r'_{i \mod n} \)). 
The angle for each distance \( s \) is calculated as:
\begin{equation}
  \theta_i^{(s)} = \arccos\left( \frac{\vec{v}_{i-s,i} \cdot \vec{v}_{i,i+s}}{\|\vec{v}_{i-s,i}\| \|\vec{v}_{i,i+s}\|} \right), \quad s = 1, 2, 3,
\end{equation}
where \( \vec{v}_{i-s,i} = r'_i - r'_{i-s} \) and \( \vec{v}_{i,i+s} = r'_{i+s} - r'_i \) are vectors to points at distance \( s \). 
These angles, expressed in radians within \( [0, \pi] \), are converted to polar coordinate space and serve as features, reflecting the geometric significance of smaller angles (e.g., corners) as potential vertices. 
The corrected coordinates \( r'_i \), image features \( \mathbf{z}_i \), and angle features \( [\theta_i^{(1)}, \theta_i^{(2)}, \theta_i^{(3)}] \) are combined as \( [\mathbf{z}'_i, r'_i, \theta_i^{(1)}, \theta_i^{(2)}, \theta_i^{(3)}] \) and input into a transformer-based vertex predictor, which outputs a probability \( \hat{p}_i \in [0, 1] \) for each point being a vertex. 

Training PST involves three loss components. The offset regression loss uses Smooth L1 Loss to supervise \( \hat{o}_i \) against ground truth offsets \( o_i = g'_i - r_i \), ensuring the corrected sequence \( R' \) aligns with the original polygon:
\begin{equation}
  \mathcal{L}_\text{off} = \sum_{i} \begin{cases} 
    0.5 (\hat{o}_i - o_i)^2 & \text{if } |\hat{o}_i - o_i| < 1, \\
    |\hat{o}_i - o_i| - 0.5 & \text{otherwise}
    \end{cases}
\end{equation}

The vertex classification loss employs Binary Cross-Entropy (BCE) Loss, comparing predicted probabilities \( \hat{p}_i \) to ground truth labels \( c_i \in \{0, 1\} \):
\begin{equation}
  \mathcal{L}_\text{vert} = -\sum_{i} \left[ c_i \log(\hat{p}_i) + (1 - c_i) \log(1 - \hat{p}_i) \right]
\end{equation}

\begin{table*} [!t]
  \centering
  \small
  \caption{Comparison of HoliTracer with state-of-the-art methods on the WHU-building, GLH-water, and VHR-road datasets.}
  \vspace{-8pt}
  \begin{threeparttable}
  \begin{tabularx}{\textwidth}{@{\extracolsep{\fill}}llccccccccc}
    \toprule
    \multirow{2}{*}{Dataset} 
      & \multirow{2}{*}{Method}
      & \multicolumn{2}{c}{Vector metrics}
      & \multicolumn{4}{c}{Instance metrics} 
      & \multicolumn{2}{c}{Semantic metrics} \\
    \cmidrule(lr){3-4} \cmidrule(lr){5-8} \cmidrule(lr){9-10}  
      &                             & $PoLiS \downarrow$ & $CIoU$
      & $AP$     & $AP_s$    & $AP_m$    & $AP_l$
      & $IoU$    & $F1$      \\
      \midrule
    \multirow{9}{*}{WHU-building}
      & TS-MTA \cite{ding2020semantic}\footnotemark[1]
      & 6.94      & 14.29
      & 36.33     & 14.34     & 65.49     & \underline{41.30}
      & 85.30     & 92.02     \\
      & LCF-ALE \cite{liu2024ultra}\footnotemark[1]
      & 7.51      & 19.09
      & 41.78     & 20.05     & 64.44     & 38.98
      & 85.67     & \underline{92.22}     \\
      & DeepSnake \cite{peng2020deep}
      & \underline{4.43}      & 53.54
      & 38.83     & 22.31     & 53.17     & 15.47
      & 76.05     & 86.01     \\
      & E2EC \cite{zhang2022e2ec}
      & 9.10      & 47.66
      & 29.85     & 17.03     & 41.32     & 8.87
      & 65.44     & 78.49     \\
      & FFL \cite{girard2021polygonal}
      & 9.34      & 35.51
      & 31.39     & 17.03      & 48.29     & 26.92
      & 71.48     & 83.25     \\
      & UniVec \cite{yang2024univecmapper}
      & 8.36      & \underline{62.26}
      & 30.13     & 16.86     & 42.18     & 27.90
      & 78.66     & 87.96     \\
      & HiSup \cite{xu2023hisup}
      & 5.49      & 42.23
      & \underline{56.77}     & \underline{36.08}     & \underline{69.44}     & 37.82
      & \underline{85.79}     & 92.21     \\
      & Ours
      & \textbf{3.63}      & \textbf{82.30}
      & \textbf{61.07}     & \textbf{40.37}     & \textbf{80.30}     & \textbf{60.00}
      & \textbf{91.60}     & \textbf{95.41}    \\
      \midrule
    \multirow{9}{*}{GLH-water}
      & TS-MTA \cite{ding2020semantic}\footnotemark[1]
      & 150.21     & 8.72
      & 2.00      & 0.50      & \underline{6.22}      & 13.62
      & 66.56     & 78.16     \\
      & LCF-ALE \cite{liu2024ultra}\footnotemark[1]
      & 159.69     & 29.40
      & 2.22      & 0.10      & 3.46      & \underline{22.16}
      & \underline{70.74}     & 80.20     \\
      & DeepSnake \cite{peng2020deep}
      & \underline{115.51}     & 43.53
      & \underline{2.54}      & \underline{1.81}      & 4.77      & 5.56
      & 64.19     & 75.42     \\
      & E2EC \cite{zhang2022e2ec}
      & 147.00     & \underline{46.50}
      & 1.71      & 1.25      & 2.43      & 7.97
      & 69.49     & 80.26     \\
      & FFL \cite{girard2021polygonal}
      & 129.45     & 41.42
      & 0.36      & 0.03      & 0.59      & 2.65
      & 67.94     & 79.58     \\
      & UniVec \cite{yang2024univecmapper}
      & 151.94     & 43.86
      & 0.77      & 0.01      & 0.31      & 9.54
      & 69.85     & \underline{80.30}     \\
      & HiSup \cite{xu2023hisup}
      & 164.56     & 16.56
      & 1.96      & 0.47      & 4.96      & 10.07
      & 49.94     & 62.91     \\
      & Ours
      & \textbf{81.87}     & \textbf{59.24}
      & \textbf{20.84}     & \textbf{19.88}     & \textbf{38.77}     & \textbf{72.29}
      & \textbf{85.68}     & \textbf{91.51}     \\
      \midrule
    \multirow{9}{*}{VHR-road}
      & TS-MTA \cite{ding2020semantic}\footnotemark[1]
      & 222.29     & 0.71
      & 0.01      & 0.01      & 0.07      & 0.01
      & 19.40     & 30.35     \\
      & LCF-ALE \cite{liu2024ultra}\footnotemark[1]
      & 249.24     & 1.32
      & 0.11      & 0.01      & 0.06      & 0.22
      & 19.87     & 30.61     \\
      & DeepSnake \cite{peng2020deep}
      & 240.12      & \underline{4.67}
      & 0.11      & 0.01      & 0.02      & 0.20
      & 18.19     & 29.13     \\
      & E2EC \cite{zhang2022e2ec}
      & 227.86      & 3.36
      & 0.09      & \underline{0.03}      & 0.08      & 0.13
      & 19.81     & 31.19     \\
      & FFL \cite{girard2021polygonal}
      & \underline{221.44}     & 1.75
      & 0.08      & 0.01      & 0.01      & 0.17
      & 21.10     & 33.04     \\
      & UniVec \cite{yang2024univecmapper}
      & 255.10      & 3.99
      & 0.01      & 0.01      & 0.05      & 0.01
      & 12.84     & 21.78     \\
      & HiSup \cite{xu2023hisup}
      & 299.03      & 3.43
      & \underline{1.51}      & 0.01      & \underline{0.14}      & \underline{0.32}
      & \underline{36.40}     & \underline{49.77}     \\
      & Ours
      & \textbf{134.13}           & \textbf{6.10}
      &\textbf{1.58}           &\textbf{0.08}          &\textbf{0.40}          &\textbf{3.99}
      & \textbf{46.48}           &\textbf{60.63}          \\
    \bottomrule
  \end{tabularx}
  \begin{tablenotes}[flushleft]
    \footnotesize
    \item[1] The method is a semantic segmentation approach. We use TC89-KCOS \cite{teh1989detection} and DP algorithm \cite{douglas1973algorithms} for vectorization.
  \end{tablenotes}
  \end{threeparttable}
  \vspace{-12pt}
  \label{tab:result}
\end{table*}

Recognizing that smaller angles indicate vertices, an angle penalty loss encourages high \( \hat{p}_i \) for points with angles below a threshold \( \theta_{\text{threshold}} \):
\begin{equation}
  \mathcal{L}_\text{angle} = \sum_{i} \begin{cases}
      \max(0, \theta_i^{(1)} - \theta_{\text{threshold}}), & \text{if } c_i = 1 \\
      \max(0, \theta_{\text{threshold}} - \theta_i^{(1)}), & \text{if } c_i = 0
    \end{cases}
  \end{equation}

The total loss is:
\begin{equation}
  \mathcal{L} = \lambda_1 \mathcal{L}_\text{off} + \lambda_2 \mathcal{L}_\text{vert} + \lambda_3 \mathcal{L}_\text{angle},
\end{equation}
where \( \lambda_1, \lambda_2, \lambda_3 \) balance the contributions of offset, classification, and angle loss. 
Under the supervision of these losses, PST refines the polygon while accurately identifying vertices based on geometric and contextual cues.

\section{Experiments}

In this section, we first introduce the datasets and evaluation metrics, followed by Section 4.2, where we compare our method with current state-of-the-art (SOTA) methods across different datasets. 
Section 4.3 presents ablation studies for the various components of our method.

\subsection{Datasets, Metrics, and Training Details}
\textbf{Datasets}. 
To evaluate HoliTracer's capability to extract different geographic entity types from large-size RSI, we conduct experiments on three large-size datasets: the WHU-building dataset~\cite{ji2018fully}, the GLH-water dataset~\cite{li2024glh}, and the VHR-road dataset, a newly constructed dataset introduced in this work.  
Each sample in these datasets exceeds a resolution of $10,000 \times 10,000$ pixels and offers ultra-high geographical resolution.

\textbf{WHU-building Dataset}.  
The WHU-building dataset is designed for precise building extraction.  
It includes a large regional image, which we partition into images of $10,000 \times 10,000$ pixels, resulting in 400 large-size RSIs with corresponding building labels.  
Each sample has a spatial resolution of 0.075 meters.  
We allocate 320 images for training, 40 for validation, and 40 for testing.

\textbf{GLH-water Dataset}.  
The GLH-water dataset is tailored for global fine-grained water body extraction, comprising 250 large-size RSIs with associated water body labels.  
Each sample measures $12,800 \times 12,800$ pixels with a spatial resolution of 0.3 meters.  
We use 200 images for training, 25 for validation, and 25 for testing.

\textbf{VHR-road Dataset}.  
The VHR-road dataset is a newly constructed dataset introduced in this work.  
Beyond road centerline extraction, large-size high-resolution RSIs enable the identification of road width attributes.  
However, existing road datasets lack suitability for large-size RSI vectorization tasks.  
To address this, we present the Very-High-Resolution Road (VHR-road) dataset, treating roads as planar elements.  
This dataset contains 208 images, each with a resolution of $12,500 \times 12,500$ pixels and a spatial resolution of 0.2 meters.  
Of these, 166 images are used for training, 21 for validation, and 21 for testing. \textit{More details are provided in the supplementary Sec.~\ref{sec:supp3}.}

\begin{figure*}
  \begin{center}
    \includegraphics[width=1\textwidth]{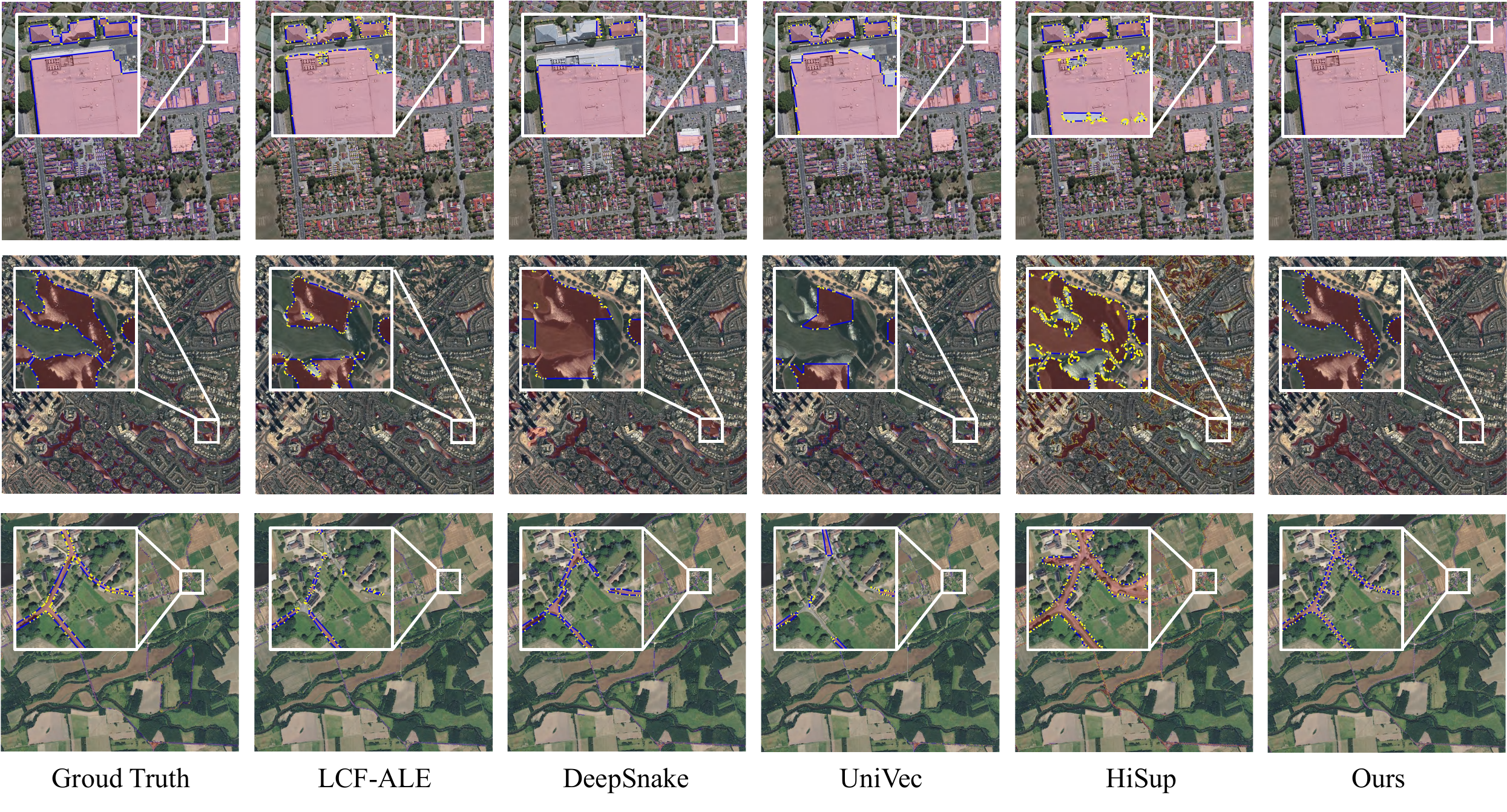}
  \end{center}
  \vspace{-12pt}
     \caption{
      Visualization of vectorization results from HoliTracer and comparative methods on WHU-building, GLH-water, and VHR-road datasets. Our method produces more accurate and complete vector representations compared to existing patch-based methods.
      }
      \vspace{-10pt}
  \label{fig:results}
\end{figure*}

\textbf{Evaluation Metrics}.  
We adopt a comprehensive set of metrics to evaluate HoliTracer and comparative methods, categorized into three levels: vector metrics, instance metrics, and semantic metrics.  
Vector metrics include $CIoU$~\cite{zorzi2022polyworld} and $PoLiS$~\cite{avbelj2014metric}.  
For instance metrics, we employ the $AP$ (average precision) metric from the COCO dataset~\cite{lin2014microsoft}.  
Additionally, we use $AP_s$, $AP_m$, and $AP_l$ to assess the accuracy of instances across different sizes, where $s$, $m$, and $l$ denote small, medium, and large instances, respectively.  
For semantic metrics, we utilize the common segmentation metrics $IoU$ and $F1$.  
\textit{Detailed calculations are provided in the supplementary Sec.~\ref{sec:supp2}.}

\textbf{Training Details}.  
For multi-scale pyramid construction, scale factors ($d$) are set to \{1, 3, 6\} for buildings and \{1, 5, 10\} for water bodies and roads.  
During MCR reconstruction, the parameter \(\epsilon\) is set to 5.  
Interpolation distances ($l$) are set to 25 for small targets (buildings) and 50 for larger targets (water bodies and roads).  
All loss function hyperparameters are set to 1, and the angle penalty \(\theta_{\text{threshold}}\) is fixed at 135 degrees.  
\textit{Further details are available in the supplementary Sec.~\ref{sec:supp3}.}

\subsection{Comparison with SOTA Methods}
To validate HoliTracer's advantages in vectorizing large-size RSI, we compare it with existing state-of-the-art (SOTA) vectorization methods, including segmentation-based methods FFL~\cite{girard2021polygonal} and HiSup~\cite{xu2023hisup}, contour-based methods DeepSnake~\cite{peng2020deep} and E2EC~\cite{zhang2022e2ec}, and the graph-based method UniVec~\cite{yang2024univecmapper}.  
Notably, when applying these methods to large-size RSI, we adopt a patch-based strategy, dividing the large-size RSI into patches and processing them independently.  
Additionally, we include large-size segmentation methods TS-MTA~\cite{ding2020semantic} and LCF-ALE~\cite{liu2024ultra}, which perform vectorization using the TC89-KCOS~\cite{teh1989detection} and DP~\cite{douglas1973algorithms} algorithms, in the comparison.

The quantitative comparison results are presented in Tab.~\ref{tab:result}, demonstrating that our method outperforms current approaches across vector, instance, and semantic metrics.  
Visual comparison results are shown in Fig.~\ref{fig:results} and \textit{supplementary Sec.~\ref{sec:supp5}}, where our method yields more accurate and complete vector representations compared to existing patch-based methods.  
The following subsections analyze performance across different datasets:

\textbf{Performance on WHU-building Dataset}.  
While existing methods achieve acceptable performance on smaller building objects within large-size RSIs, they struggle to extract larger buildings that require information from a local-to-global attention mechanism.  
HoliTracer's superior perception with a local-to-global attention mechanism enables it to accurately predict these challenging samples, as evidenced in Fig.~\ref{fig:motivation} and Fig.~\ref{fig:results}.  
Moreover, thanks to HoliTracer's robust vertex refinement and identification capabilities, it delivers more precise building vector representations.  
Consequently, HoliTracer outperforms the second-best segmentation method by 5.81\% on the semantic metric $IoU$, surpasses the second-best instance segmentation method by 4.30\% on the COCO instance metric $AP$ (with an 18.70\% improvement on large instances), and exceeds the second-best building vectorization method by 20.04\% on the vector metric $CIoU$.

\begin{table*} 
  \caption{The ablation studies on the PST.}
  \vspace{-8pt}
  \small
  \begin{threeparttable}
  \begin{tabularx}{\textwidth}{@{\extracolsep{\fill}}llccccccccc}
    \toprule
    \multirow{2}{*}{Dataset} 
      & \multirow{2}{*}{Vectorize method}
      & \multicolumn{2}{c}{Vector metrics}
      & \multicolumn{4}{c}{Instance metrics} 
      & \multicolumn{2}{c}{Semantic metrics}\\ 
    \cmidrule(lr){3-4} \cmidrule(lr){5-8} \cmidrule(lr){9-10}  
      &                             & $PoLiS \downarrow$ & $CIoU$
      & $AP$     & $AP_s$    & $AP_m$    & $AP_l$
      & $IoU$    & $F1$      \\
      \midrule
    \multirow{3}{*}{WHU-Building}
      & Baseline\footnotemark[1]
      & 3.83      & 18.47
      & 58.75     & 37.87     & 79.85     & 49.13
      & 91.55     & \textbf{95.54}    \\
      & Baseline\footnotemark[1] + DP \cite{douglas1973algorithms}
      & 4.02      & 60.32
      & 58.42     & 37.50     & 79.76     & 49.13
      & 91.50     & 95.52    \\
      & Baseline\footnotemark[1] + PST
      & \textbf{3.63}      & \textbf{82.30}
      & \textbf{61.07}     & \textbf{40.37}     & \textbf{80.30}     & \textbf{60.00}
      & \textbf{91.60}     & 95.41    \\
      \bottomrule
  \end{tabularx}
  \begin{tablenotes}[flushleft]
    \footnotesize
    \item[1] Using TC89-KCOS \cite{teh1989detection} to extract polygon contours.
  \end{tablenotes}
\end{threeparttable}
  \label{tab:ablation2}
\end{table*}

\begin{table*} 
  \vspace{-8pt}
  \caption{The ablation studies on the loss function and angle features.}
  \vspace{-8pt}
  \small
  \begin{threeparttable}
  \begin{tabularx}{\textwidth}{@{\extracolsep{\fill}}cccccccccccc}
    \toprule
      \multirow{2}{*}{Angle features}
      & \multirow{2}{*}{Angle loss}
      & \multirow{2}{*}{\( \theta_{\text{threshold}} \)}
      & \multicolumn{2}{c}{Vector metrics}
      & \multicolumn{4}{c}{Instance metrics}
      & \multicolumn{2}{c}{Semantic metrics}\\ 
    \cmidrule(lr){4-5} \cmidrule(lr){6-9} \cmidrule(lr){10-11}  
      &   &  & $PoLiS \downarrow$ & $CIoU$
      & $AP$     & $AP_s$    & $AP_m$    & $AP_l$
      & $IoU$    & $F1$      \\
      \midrule
      - & - & -                  
      & 3.91      & 78.27
      & 59.73     & 38.81     & 79.90     & 59.63
      & 91.47     & 95.32    \\
       \checkmark & - & -                  
      & 3.82      & 79.66
      & 60.11     & 39.37     & 80.01     & 59.66
      & 91.48     & 95.40    \\
       \checkmark & \checkmark & 90                         
       & 3.72      & 78.27
       & 60.68     & 39.63     & 80.25     & 59.20
       & 90.62     & 95.22    \\
       \checkmark & \checkmark & 135                         
      & \textbf{3.63}      & \textbf{82.30}
      & \textbf{61.07}     & \textbf{40.37}     & \textbf{80.30}     & \textbf{60.00}
      & \textbf{91.60}     & \textbf{95.41}    \\
       \checkmark & \checkmark & 180                         
      & 3.72      & 81.28
       & 60.76     & 39.85     & 80.21     & 59.76
       & 91.34     & 95.41    \\
       \bottomrule
  \end{tabularx}
 \end{threeparttable}
  \label{tab:ablation3}
 \end{table*}

 \textbf{Performance on GLH-water Dataset}.  
 Compared to buildings, water bodies exhibit greater scale variations, amplifying our method's advantages.  
 Specifically, HoliTracer outperforms the second-best methods by 14.94\%, 18.30\%, and 12.74\% on the $IoU$, $AP$, and $CIoU$ metrics, respectively.  
 This superior performance stems from the severe large-size challenges encountered by competing methods when processing large-scale water body targets, as illustrated in Fig.~\ref{fig:results}.
 
 \textbf{Performance on VHR-road Dataset}.  
 Unlike buildings and water bodies, roads typically consist of a single connected target per sample, rendering instance metrics less relevant for evaluation.  
 Nevertheless, our method outperforms the second-best methods by 10.04\%, 3.67\%, and 1.43\% on the $IoU$, $AP_l$, and $CIoU$ metrics, respectively.  
 As shown in Fig.~\ref{fig:results}, HoliTracer produces roads with stronger connectivity and reduced fragmentation.  
 Although HoliTracer achieves the best performance on the VHR-road dataset, road vectorization remains challenging due to the complexity of road network structures.

\begin{table}  
  \vspace{-10pt}
  \caption{The ablation studies on the image pyramid within CAN.}
  \vspace{-6pt}
  \centering  
  \small
  \begin{tabularx}{0.45\textwidth}{@{\extracolsep{\fill}}llccc} 
    \toprule
    Dataset                                                     & Context    & IoU           & F1            \\
    \midrule
    \multirow{4}{*}{WHU-building}           & 1          & 91.23         & 95.29         \\
                                                              & 1, 3, 6      & \textbf{92.21} & \textbf{95.94} \\
                                                              & 1, 4, 8      & 92.12         & 95.68         \\
                                                              & 1, 5, 10     & 92.14         & 95.72         \\
    \bottomrule
  \end{tabularx}
  \label{tab:ablation1}
\end{table}

\subsection{Ablation Studies}

\begin{figure}
  \begin{center}
    \includegraphics[width=0.48\textwidth]{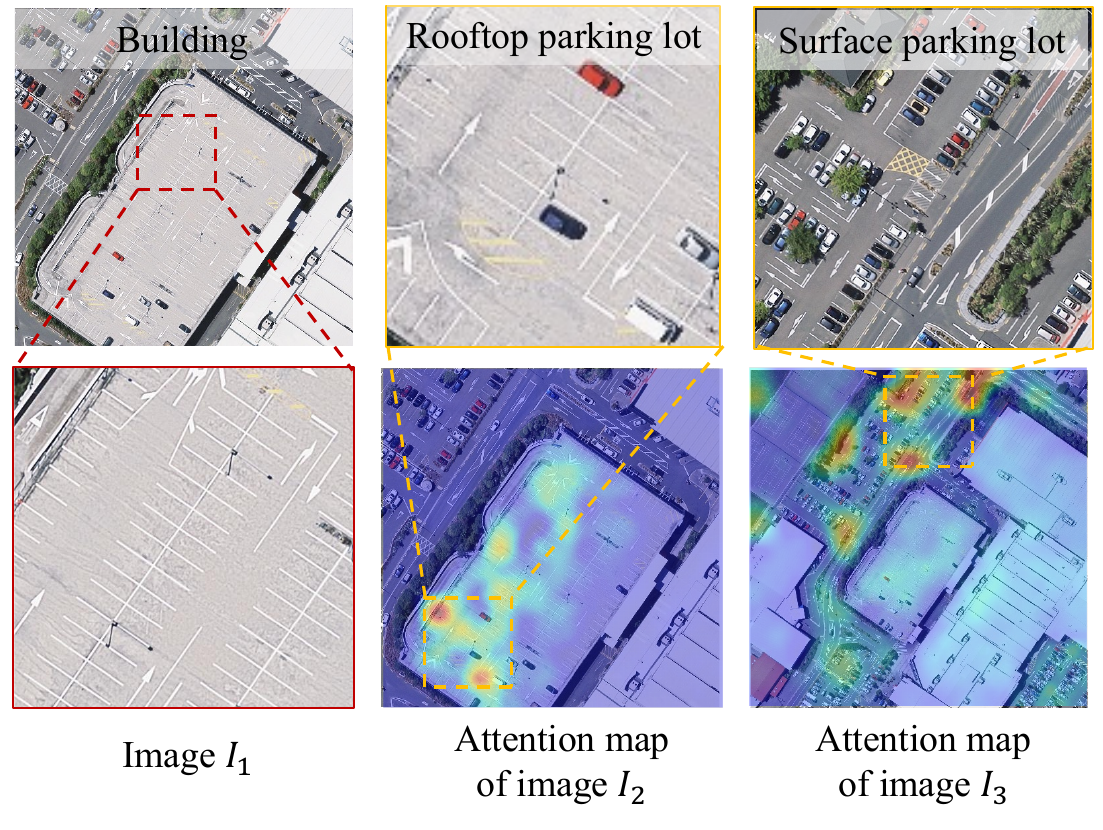}
  \end{center}
  \vspace{-16pt}
     \caption{The attention map of the CAN at different scales. Image $I_1$ shows a rooftop parking lot on a building. In the attention map of Image $I_2$, CAN focuses on the context of the building, while in the attention map of Image $I_3$, CAN focuses on the features of the surface parking lot.}
  \vspace{-16pt} 
     \label{fig:attention}
\end{figure}

\textbf{Ablation of Image Pyramid}.  
We conduct an ablation study on the image pyramid within CAN, evaluating performance with different downsampling ratios $d$.  
As shown in Tab.~\ref{tab:ablation1}, the results demonstrate that information from a local-to-global attention mechanism in the pyramid enhances segmentation performance.  
Since buildings are typically smaller objects, the downsampling ratios of 1, 3, and 6 produced the best results on the building dataset.  
Additionally, we visualize the attention maps of CAN in Fig.~\ref{fig:attention}.  
These visualizations reveal that CAN effectively captures information using a local-to-global attention mechanism, thereby improving the completeness of segmentations.

\textbf{Ablation of PST}.  
As presented in Tab.~\ref{tab:ablation2}, incorporating PST improves both instance and vector metrics compared to the original polygon contour results and those simplified by the DP algorithm, with particularly significant gains in the vector metric.  
Notably, while the original polygon results retain more detailed edge representations, introducing PST slightly reduces the semantic metric $F1$.  
However, as illustrated in \textit{supplementary Fig.~\ref{fig:sup_pst}}, despite the higher $F1$ of the original edge point results, they are markedly inferior to PST results as vector representations.  
\textit{Additional ablation studies are available in the supplementary Sec.~\ref{sec:supp5}.}

\textbf{Ablation of Loss Function and Angle Features}.  
We perform ablation studies on the loss function and angle features, with results shown in Tab.~\ref{tab:ablation3}.  
Excluding angle features and angle loss significantly degrades PST's performance, as the model struggles to learn explicit rules in an additional dimension.  
Furthermore, an angle threshold of 135 degrees achieves optimal performance, as excessively high or low thresholds lead to retaining too many or too few vertices, respectively.
\section{Conclusion}

In this paper, we introduce HoliTracer, the first framework tailored for holistic vector map extraction from large-size RSI.  
Within HoliTracer, we enhance segmentation of large-size RSI using the CAN, which employs a local-to-global attention mechanism to capture contextual information.
Furthermore, we achieve holistic vectorization through a robust pipeline that employs the MCR to reconstruct polygons and the PST to trace essential vertices.  
Extensive experiments on multiple datasets (including buildings, water bodies, and roads) confirm that HoliTracer outperforms SOTA approaches, providing an effective solution for real-world vector mapping from large-size RSI.

\section*{Acknowledgments}
This work was supported by the National Natural Science Foundation of China under Grants 42371321 and 424B2006.

{
    \small
    \bibliographystyle{ieeenat_fullname}
    \bibliography{main}
}

\clearpage
\setcounter{page}{1}
\maketitlesupplementary

\section{MCR Algorithm Details}
The MCR algorithm is detailed in Algorithm \ref{alg:mcr}, which is used to reconstruct polygon contours and align them with ground truth polygons.
\begin{algorithm}
  \caption{Mask Contour Reformer}
  \label{alg:mcr}
  \begin{algorithmic}[1]
  \State \textbf{Input:} \( S \), \( G = [g_1, g_2, \dots, g_M] \), \( \epsilon \), \( l \)
  \State \textbf{Output:} \( R \), \( G' \), \( C \)
  \State \( S' = [s'_1, s'_2, \dots, s'_m] \) \Comment{DP simplification with \( \epsilon \)}
  \State \textbf{For} each \( (s'_i, s'_{i+1}) \) in \( S' \) (with \( s'_{m+1} = s'_1 \) if closed):
  \State \quad \( \vec{v}_i = s'_{i+1} - s'_i \), \( q_i = \|\vec{v}_i\| \), \( \hat{v}_i = \vec{v}_i / q_i \)
  \State \quad \( K_i = \lfloor q_i / l \rfloor \)
  \State \quad \( [s'_i + k d \hat{v}_i \text{ for } k = 0 \text{ to } K_i] + [s'_{i+1}] \)
  \State \( R = [r_1, r_2, \dots, r_N] \) \Comment{Concatenate all points}
  \State \textbf{For} each \( g_j \in G \):
  \State \quad \( r_{k_j} = \arg\min_{r \in R} \| r - g_j \| \)
  \State Vertices: \( \{ r_{k_j} | j = 1, 2, \dots, M \} \)
  \State Indices: \( i_1 < i_2 < \dots < i_M \) in \( R \)
  \State \textbf{For} each \( k = 1 \) to \( M \) (with \( i_{M+1} = i_1 \) if closed):
  \State \quad \( n_k = i_{k+1} - i_k - 1 \)
  \State \quad \( p_{k,m} = g_k + \frac{m}{n_k + 1} (g_{k+1} - g_k) \) for \( m = 1 \) to \( n_k \)
  \State \( G' = [g_1, p_{1,1}, \dots, p_{1,n_1}, g_2, \dots, g_M, p_{M,1}, \dots, p_{M,n_M}] \)
  \State \( C = [c_1, \dots, c_N] \), \( c_i = 1 \) if \( r_i \) is a vertex, else \( 0 \)
  \State \textbf{Return} \( R, G', C \)
  \end{algorithmic}
  \end{algorithm}

\section{Metrics for Evaluation} \label{sec:supp2}
For a comprehensive assessment of semantic segmentation, instance segmentation, and vector generation quality, we report three widely used categories of metrics: semantic metrics, instance metrics, and vector metrics.

\textbf{Vector Metrics}.  
Vector metrics include PoLiS~\cite{avbelj2014metric} and Complexity-aware IoU (C-IoU)~\cite{zorzi2022polyworld}.  
For two given polygons $A$ and $B$, PoLiS is defined as the average distance between each vertex $a_j \in A$, $j=1, \ldots, q$, of $A$ and its closest point ($b$) on the boundary $\partial B$, and vice versa.  
Assuming polygon $B$ has vertices $b_k \in B$, $k=1, \ldots, r$, the PoLiS metric~\cite{avbelj2014metric} is expressed as:
\begin{equation}
\begin{aligned}
{\rm PoLiS}(A, B) = & \frac{1}{2q} \sum_{a_j \in A} \min_{b \in \partial B} \|a_j - b\| \\
    & + \frac{1}{2r} \sum_{b_k \in B} \min_{a \in \partial A} \|b_k - a\| \,,
\end{aligned}
\end{equation}
where $\frac{1}{2q}$ and $\frac{1}{2r}$ are normalization factors.  
The IoU threshold for filtering predicted building polygons is set to 0.5, following~\cite{xu2023hisup}.  
A lower PoLiS value indicates greater similarity between predicted and ground truth polygons.  
The Complexity-aware IoU (C-IoU)~\cite{zorzi2022polyworld} is also computed for polygon evaluation, defined as:
\begin{equation}
\text{C-IoU}(A, B) = {\rm IoU}(A_m, B_m) \cdot (1 - {\rm RD}(N_A, N_B)) \,,
\end{equation}
where ${\rm IoU}(A_m, B_m)$ denotes the standard IoU between the polygon masks $A_m$ and $B_m$, and ${\rm RD}(N_A, N_B) = |N_A - N_B| / (N_A + N_B)$ represents the relative difference between the number of vertices $N_A$ in polygon $A$ and $N_B$ in polygon $B$.  
C-IoU balances segmentation and polygonization accuracy while penalizing both oversimplified and overly complex polygons relative to the ground truth complexity.

For the road dataset, we also report the Average Path Length Similarity (APLS) metric~\cite{van2018spacenet}, which measures road network similarity by comparing the path lengths between node pairs in the predicted and ground truth graphs.
The APLS metric is defined as follows. Given a ground-truth graph $G_{gt}$ and a predicted graph $G_{pred}$, the APLS is computed based on the symmetric difference of shortest path lengths between all reachable pairs of nodes in both graphs:

\[
\mathrm{APLS} = 1 - \frac{1}{|P|} \sum_{(i, j) \in P} \frac{|d_{pred}(i,j) - d_{gt}(i,j)|}{d_{gt}(i,j)}
\]
where $P$ is the set of all node pairs $(i,j)$ with valid paths in $G_{gt}$, $d_{gt}(i,j)$ is the shortest path length between nodes $i$ and $j$ in the ground truth graph, and $d_{pred}(i,j)$ is the corresponding path in the predicted graph. The APLS value ranges from 0 to 1, with higher values indicating better topological alignment.

\begin{table*}[h]
  \centering
  \small
  \caption{Comparison with segmentation method on WHU-Building.}
  \label{tab:segmentation}
  \begin{tabular}{@{}lcccccccc@{}}
    \toprule
    Method & PoLiS $\downarrow$ & CIoU & AP & APs & APm & APl & IoU & F1 \\
    \midrule
    HRNet + DP   & 6.10 & 50.01 & 48.13 & 26.58 & 70.59 & 38.60 & 86.51 & 92.68 \\
    HRNet + PST  & 5.91 & 61.51 & 49.01 & 26.59 & 71.30 & 50.74 & 86.48 & 92.67 \\
    CAN + DP     & 4.02 & 60.32 & 58.42 & 37.50 & 79.76 & 49.13 & 91.50 & \textbf{95.52} \\
    CAN + PST    & \textbf{3.63} & \textbf{82.30} & \textbf{61.07} & \textbf{40.37} & \textbf{80.30} & \textbf{60.00} & \textbf{91.60} & 95.41 \\
    \bottomrule
  \end{tabular}
\end{table*}

\begin{table*}[h] 
  \small
  \centering
  \caption{APLS metric comparison with various vectorization methods.}
  \label{tab:apls} 
  \begin{tabular}{@{}lcccccccc@{}}
    \toprule
    Metric & TS-MTA & LCF-ALE & DeepSnake & E2EC & FFL & UniVec & HiSup & Ours \\
    \midrule
    APLS & 14.49 & 15.67 & 11.26 & 11.59 & 12.68 & 10.22 & 20.46 & \textbf{28.35} \\
    \bottomrule
  \end{tabular}
\end{table*}

\begin{table*}[h]
  \centering
  \small
  \caption{Computational cost analysis on WHU-Building}
  \vspace{-8pt}
  \label{tab:compute}
  \begin{tabular}{@{}lcccc@{}}
    \toprule
    Method & Training Time (h) & Params (M) & GPU Memory (G) & Infer Time (s/sample) \\
    \midrule
    HiSup \cite{xu2023hisup}       & 10.86 & 74.29  & 0.48 & 2568.89 \\
    Ours CAN    & 16.27 & 268.52 & 1.88 & 125.31  \\
    Ours MCR    & -     & -      & -    & 132.78  \\
    Ours PST    & 15.66 & 43.24  & 1.25 & 200.01  \\
    Ours (ALL)  & 31.93 & 311.76 & 1.88 & 458.10  \\
    \bottomrule
  \end{tabular}
\end{table*}

\textbf{Instance Metrics}.  
Instance metrics adopt the standard COCO measure, mean Average Precision (AP), calculated over multiple Intersection over Union (IoU) thresholds.  
AP is averaged across ten IoU values ranging from 0.50 to 0.95 with a step size of 0.05, rewarding detectors with better localization.  
Additionally, $AP_{(S,M,L)}$ is used to evaluate performance on objects of different sizes.  
Given that geographical instances occupy more pixels in large-size very-high-resolution (VHR) remote sensing images, we redefine size categories relative to the COCO standard: small, medium, and large correspond to areas $< 128^2$, between $128^2$ and $512^2$, and $> 512^2$ pixels, respectively, where the area is measured as the number of pixels in the segmentation mask.

\textbf{Semantic Metrics}.  
Semantic metrics include the F1-score and Mean Intersection over Union (MIoU).  
The F1-score, the harmonic mean of Precision and Recall, provides a comprehensive evaluation of model performance.  
MIoU, the average ratio of intersection to union between predicted and ground truth segmentation results, reflects the model’s overall segmentation effectiveness across the entire image.

\section{Implementation Details} \label{sec:supp3}

\textbf{Details of VHR-road Dataset}.  
The VHR-road dataset is comprised of high-resolution remote sensing imagery of major urban areas in France, acquired from BD ORTHO \cite{BDOrtho}. 
The corresponding raw road labels are sourced from European Union's Copernicus Land Monitoring Service information \cite{UrbanAtlas2018}.
We subsequently filter and rectify inaccuracies within these labels, culminating in a final dataset of 208 image tiles, each with a dimension of $12500 \times 12500$ pixels.

\textbf{Hyperparameter Settings}.  
When constructing the multi-scale pyramid, we set the scale factors $d$ to \{1, 3, 6\} for buildings and \{1, 5, 10\} for water bodies and roads.  
During boundary point reconstruction, the Douglas-Peucker simplification parameter \(\epsilon\) is set to 5.  
For small targets such as buildings, the interpolation distance $l$ is set to 25, while for larger targets like water bodies and roads, it is set to 50.

\textbf{Training Process}.  
HoliTracer involves two training processes.  
For CAN training, we employ the Adam optimizer with a learning rate of 0.0001.  
For PST training, we use the Adam optimizer with a learning rate of 0.01.  
All loss function hyperparameters are set to 1, and the angle penalty term \(\theta_{\text{threshold}}\) is fixed at 135 degrees.  
All experiments are conducted using the PyTorch framework on four NVIDIA A100 GPUs.

\section{Supplementary Experiments}

\textbf{Comparison with Segmentation Methods and Flexibility of PST}.
Table~\ref{tab:segmentation} presents direct comparisons with the segmentation-based method HRNet \cite{wang2020deep}, showing the superior performance of our Context Attention Net (CAN). 
Although the main focus is on vectorization methods, segmentation-based approaches are also considered, such as Hisup, which uses HRNet as its backbone. 
To evaluate the flexibility of the proposed PST, the table also reports results for HRNet+PST and HRNet+DP simplification. 
PST consistently achieves better performance in vectorizing general segmentation masks. 
While CAN enhances PST’s performance on large-size RSI within the complete HoliTracer pipeline, these results demonstrate PST’s general utility across different segmentation outputs.

\textbf{Evaluation with APLS Metric.}
To further evaluate the quality of road network vectorization, the Average Path Length Similarity (APLS) metric is adopted. 
Table~\ref{tab:apls} reports the APLS scores for different methods on the road dataset.

\textbf{Computational Efficiency and Scalability.}
Table~\ref{tab:compute} summarizes the computational complexity and inference performance of HoliTracer. 
Although HoliTracer has more parameters and a higher computational load compared to lightweight methods, the overhead remains acceptable for vectorization tasks, which typically do not require real-time processing.
HoliTracer processes large images directly, eliminating the need for patch-wise inference and subsequent stitching. 
This design reduces total inference time. 
Scalability experiments confirm that HoliTracer handles images up to $40{,}000 \times 50{,}000$ pixels using 64GB of CPU RAM. 
Our implementation also includes GPU-based parallelism for inference on large images, providing robust scalability across different computational environments.
\section{Supplementary Ablation Studies} \label{sec:supp5}
\begin{table*} 
  \caption{The ablation studies on the PST.}
  \vspace{-8pt}
  \small
  \begin{threeparttable}
  \begin{tabularx}{\textwidth}{@{\extracolsep{\fill}}llccccccccc}
    \toprule
    \multirow{2}{*}{Dataset} 
      & \multirow{2}{*}{Vectorize method}
      & \multicolumn{2}{c}{Vector metrics}
      & \multicolumn{4}{c}{Instance metrics} 
      & \multicolumn{2}{c}{Semantic metrics}\\ 
    \cmidrule(lr){3-4} \cmidrule(lr){5-8} \cmidrule(lr){9-10}  
      &                             & $PoLiS \downarrow$ & $CIoU$
      & $AP$     & $AP_s$    & $AP_m$    & $AP_l$
      & $IoU$    & $F1$      \\
      \midrule
    \multirow{3}{*}{WHU-Building}
      & Baseline\footnotemark[1]
      & 3.83      & 18.47
      & 58.75     & 37.87     & 79.85     & 49.13
      & 91.55     & \textbf{95.54}    \\
      & Baseline\footnotemark[1] + DP \cite{douglas1973algorithms}
      & 4.02      & 60.32
      & 58.42     & 37.50     & 79.76     & 49.13
      & 91.50     & 95.52    \\
      & Baseline\footnotemark[1] + PST
      & \textbf{3.63}      & \textbf{82.30}
      & \textbf{61.07}     & \textbf{40.37}     & \textbf{80.30}     & \textbf{60.00}
      & \textbf{91.60}     & 95.41    \\
      \midrule
    \multirow{3}{*}{GLHWater}
      & Baseline\footnotemark[1]
      & 82.85     & 26.74
      & 20.30     & 10.48     & 35.44     & 58.25
      & \textbf{85.76}     & \textbf{91.55}    \\
      & Baseline\footnotemark[1] + DP \cite{douglas1973algorithms}
      & 83.72     & 55.08
      & 20.31     & \textbf{10.53}     & 35.34     & 58.25
      & 85.74     & 91.54    \\
      & Baseline\footnotemark[1] + PST
      & \textbf{82.42}     & \textbf{57.88}
      & \textbf{20.84}     & 10.08     & \textbf{38.08}     & \textbf{70.35}
      & 85.50     & 91.40    \\
      \midrule
    \multirow{3}{*}{VHRRoad}
      & Baseline\footnotemark[1]
      & 135.05    & 1.43
      & \textbf{1.71}      & 0.08      & \textbf{0.43}      & 3.74
      & \textbf{46.80}     & \textbf{61.03}    \\
      & Baseline\footnotemark[1] + DP \cite{douglas1973algorithms}
      & 138.01    & 5.41
      & 1.70      & 0.08      & 0.42      & 3.74
      & 46.65     & 60.88    \\
      & Baseline\footnotemark[1] + PST
      & \textbf{134.13}           & \textbf{6.10}
      &1.58           &\textbf{0.08}          &0.40          &\textbf{3.99}
       &46.48           &60.63          \\
      \bottomrule
  \end{tabularx}
  \begin{tablenotes}[flushleft]
    \footnotesize
    \item[1] Using TC89-KCOS \cite{teh1989detection} to extract polygon contours.
  \end{tablenotes}
\end{threeparttable}
  \label{tab:ablation5}
\end{table*}

\begin{table}  
  \caption{The ablation studies on the image pyramid within CAN.}
  \vspace{-8pt}
  \centering  
  \small
  \begin{tabularx}{0.45\textwidth}{@{\extracolsep{\fill}}llccc} 
    \toprule
    Dataset                                                     & Context    & IoU           & F1            \\
    \midrule
    \multirow{4}{*}{WHU-building}           & 1          & 91.23         & 95.29         \\
                                                              & 1, 3, 6      & \textbf{92.21} & \textbf{95.94} \\
                                                              & 1, 4, 8      & 92.12         & 95.68         \\
                                                              & 1, 5, 10     & 92.14         & 95.72         \\
    \midrule
    \multirow{4}{*}{GLH-water}                & 1          & 85.73         & 91.14         \\
                                                              & 1, 3, 6      & 86.45         & 92.59   \\
                                                              & 1, 4, 8      & 86.90         & 93.00              \\
                                                              & 1, 5, 10     & \textbf{87.95} & \textbf{93.59} \\
    \midrule
    \multirow{4}{*}{VHR-road}                & 1          & 48.35         & 65.09              \\
                                                              & 1, 3, 6      & 49.74         & 66.32     \\
                                                              & 1, 4, 8      & 49.69         & 66.40               \\
                                                              & 1, 5, 10     & \textbf{50.47} & \textbf{67.09}  \\
    \bottomrule
  \end{tabularx}
  \label{tab:ablation4}
\end{table}

To further investigate the effectiveness of the image pyramid within CAN and the PST, we conduct additional ablation studies on the other two datasets.  
This also serves to justify our choice of different hyperparameter settings across datasets.

Table~\ref{tab:ablation4} reports ablation studies on the image pyramid within CAN.  
We compare different scale settings on three datasets.  
The results show that scales \{1, 3, 6\} work best for buildings, while \{1, 5, 10\} perform best for water bodies and roads.  
This indicates that smaller buildings need less context, and too much context may harm performance.  
In contrast, larger and more connected water bodies and roads benefit from more contextual information.  
Note that segmentation is evaluated at the pixel level, so semantic metrics differ from vectorized semantic metrics.

Table~\ref{tab:ablation5} presents ablation studies on the PST.  
Comparing the baseline and PST methods across three datasets, PST notably improves vector and instance metrics.  
Although semantic metrics slightly decrease, PST achieves better instance extraction and vector representation.

\section{Supplementary Visualization Results}\label{sec:supp5}
\begin{figure}
  \begin{center}
    \includegraphics[width=0.47\textwidth]{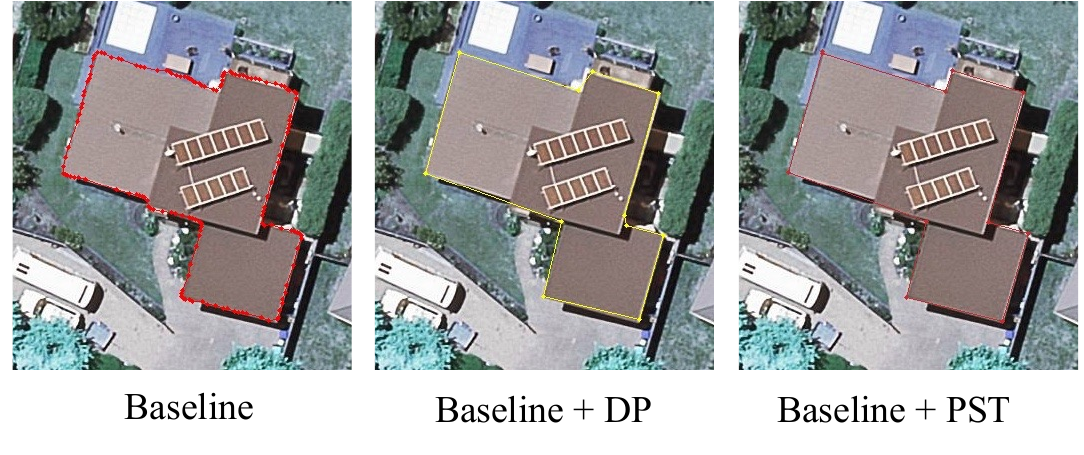}
  \end{center}
  \vspace{-16pt}
     \caption{The visualization of output polygons of different methods.}
  \vspace{-12pt}
     \label{fig:sup_pst}
\end{figure}

\begin{figure*}
  \begin{center}
    \includegraphics[width=0.995\textwidth]{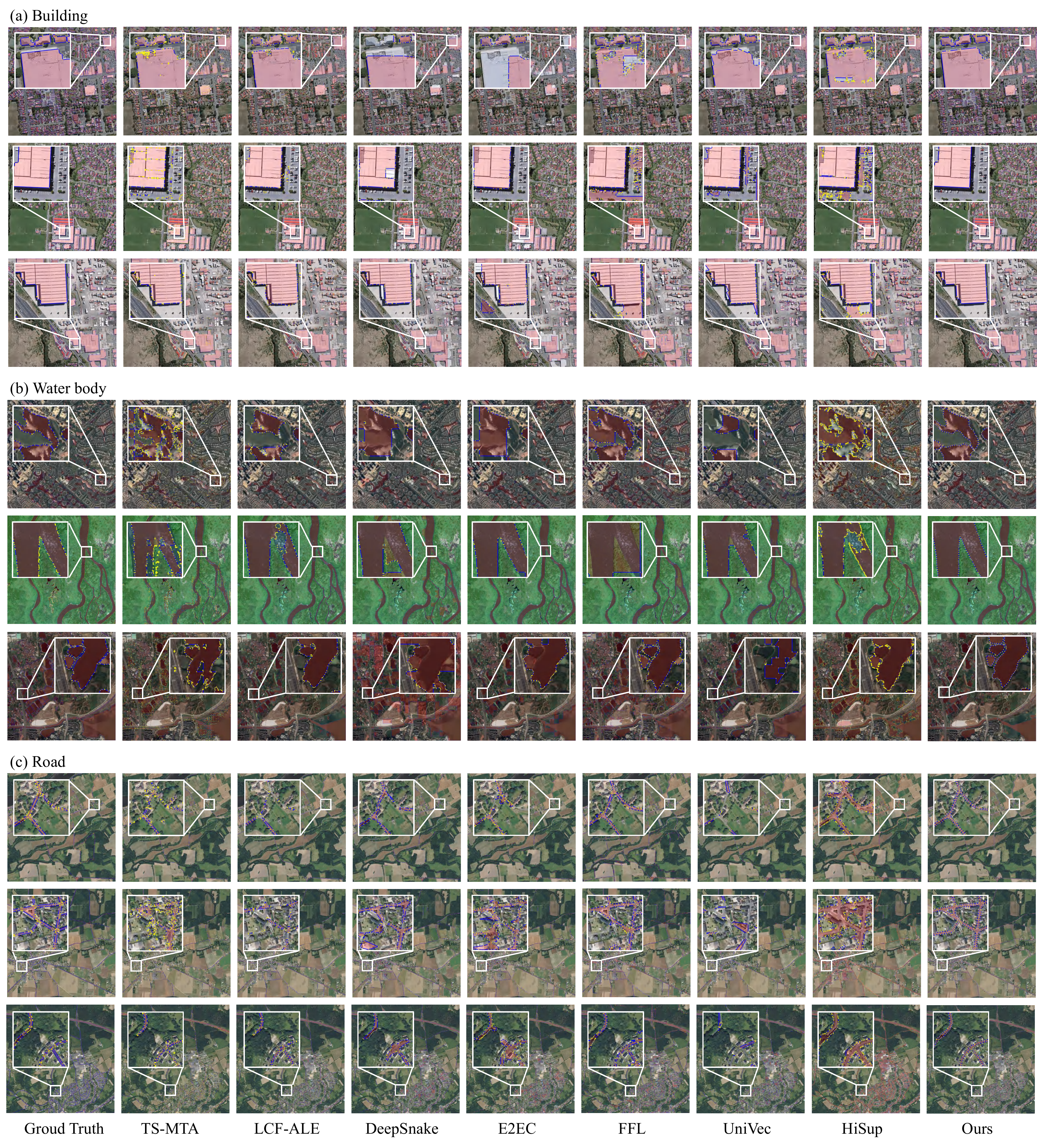}
  \end{center}
  \vspace{-12pt}
     \caption{
      Visualization of vectorization results of the all methods on WHU-building, GLH-water, and VHR-road test datasets. Our method produces more accurate and complete vector representations compared to existing patch-based methods.
      }
  \label{fig:sup_results}
\end{figure*}

\begin{figure*}
  \begin{center}
    \includegraphics[width=1\textwidth]{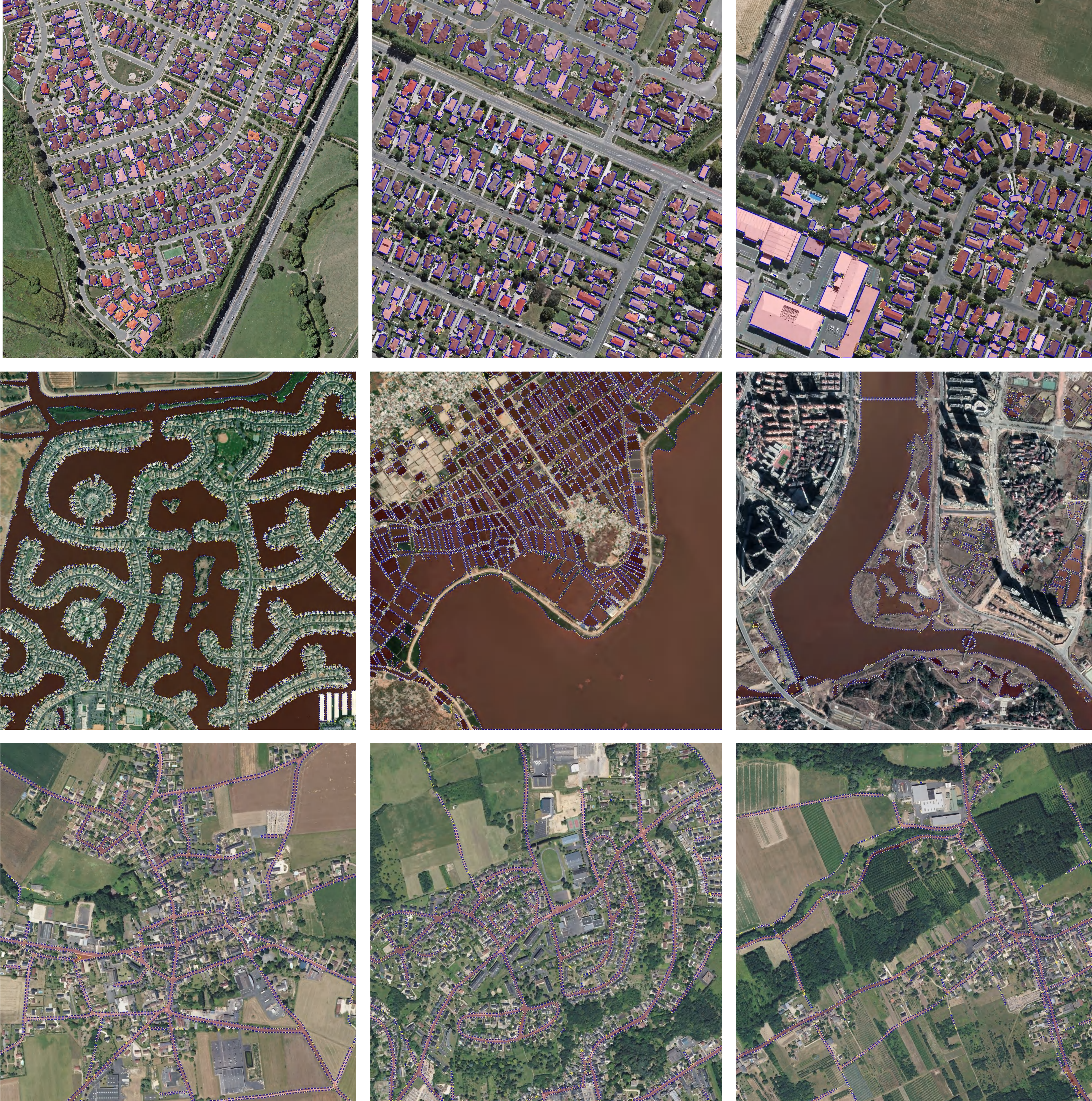}
  \end{center}
  \vspace{-12pt}
     \caption{
      Visualization of HoliTracer's vectorization results on large-size RSI.
      }
  \label{fig:sup_results2}
\end{figure*}

We provide additional visualization results on the WHU-building, GLH-water, and VHR-road datasets in Fig.~\ref{fig:sup_pst}, Fig.~\ref{fig:sup_results} and Fig.~\ref{fig:sup_results2}.  
Fig.~\ref{fig:sup_pst} illustrates the output polygons of different methods, including the baseline method using TC89-KCOS for polygon contour extraction, baseline with DP simplification, and baseline with our proposed PST.
Fig.~\ref{fig:sup_results} displays the vectorization results of all baseline methods and our HoliTracer on the WHU-building, GLH-water, and VHR-road datasets.  
Fig.~\ref{fig:sup_results2} presents the vectorization results of HoliTracer on large-size RSI.  
The results demonstrate that HoliTracer produces more accurate and complete vector representations compared to existing patch-based methods, and it effectively handles diverse geographic objects across large-size RSI.


\end{document}